\documentclass{llncs}

\usepackage{pgfplots}
\usepackage{amsmath,amsfonts,amssymb}
\usepackage{subcaption}
\begin{document}
\title{Fast Metric Learning For Deep Neural Networks}
\author{Henry Gouk\inst{1} \and Bernhard Pfahringer\inst{1} \and Michael Cree\inst{2}}
\institute{Department of Computer Science, University of Waikato, Hamilton, New Zealand\\
\email{hgrg1@students.waikato.ac.nz}, \email{bernhard@waikato.ac.nz}
\and School of Engineering, University of Waikato, Hamilton, New Zealand\\
\email{cree@waikato.ac.nz}
}
\maketitle

\begin{abstract}
Similarity metrics are a core component of many information retrieval and machine learning systems. In this work we propose a method capable of learning a similarity metric from data equipped with a binary relation. By considering only the similarity constraints, and initially ignoring the features, we are able to learn target vectors for each instance using one of several appropriately designed loss functions. A regression model can then be constructed that maps novel feature vectors to the same target vector space, resulting in a feature extractor that computes vectors for which a predefined metric is a meaningful measure of similarity. We present results on both multiclass and multi-label classification datasets that demonstrate considerably faster convergence, as well as higher accuracy on the majority of the intrinsic evaluation tasks and all extrinsic evaluation tasks.
\end{abstract}

\section{Introduction}
Many machine learning and information retrieval systems rely on distance metrics that can accurately estimate the semantic similarity between two objects of interest. Functions such as Euclidean distance and cosine similarity when applied directly to complex input domains, like images or audio, are poor measures of semantic relatedness. Hence, machine learning has been applied to the problem of constructing metrics suited for specific domains.

Recent advances in this field have taken advantage of the superior modelling capacity of deep neural networks to learn complex relations between data objects. While methods based on deep learning are able to represent more intricate metrics compared to linear Mahalanobis methods, they also take considerably longer to train. The objective of this work is to find a method to accelerate the training of deep learning approaches to similarity metric learning.

Current metric learning algorithms based on deep neural networks rely on propagating multiple training examples through the network simultaneously in order to compute the gradient of some objective functions defined over the network output. For example, one could attempt to maximise the Euclidean distance between the network outputs for dissimilar instances, while also trying to minimise the Euclidean distance between the computed representations for instances that are similar. This forces one to train the network on pairs of examples---something that greatly increases the effective training set size, and therefore the time taken for the network to converge towards a good solution.

Many techniques are evaluated on classification benchmark datasets, where each instance is labelled with a class and two instances are considered similar if and only if they share the same class. By evaluating metrics in this way there is an implicit assumption that transitivity is desired---which for some tasks is justified---but this is not always the case. For example, consider three images: A, B, and C. It could be the case the that A and B are considered similar because they have similar backgrounds. It could also be the case that B and C are similar because the same object appears in the foreground of both, despite the rest of the scene being vastly different. If transitivity were to be enforced then A and C would be considered similar, even though there is no reason for that to be the case.

We are interested in both transitive and non-transitive relations, such as multi-label classification datasets. Elaborating on the previous example; images A and B would be given a label indicating what type of background they have, while B and C would be given a label indicating what foreground object appears in the image. In this case B has two labels, and hence the task fits into the multi-label classification paradigm. However, in this work we take an even more general view of the problem definition. We assume only that the information provided consists of pairwise similarity or dissimilarity constraints. This generalised view enables the use of more diverse data collection strategies, such as the relevance feedback methods commonly used in information retrieval systems.

In our approach to similarity metric learning we acknowledge that there are latent classes within the data, however no explicit knowledge of these classes is required. By taking advantages of the existence of these latent classes, we first learn the structure of a target vector space for an embedding function, and subsequently learn a model that performs the embedding. The compute intensive part of our algorithm does not operate on pairs of feature vectors, and hence results in a computationally cheaper approach to learning instance similarity.

We first review some related methods in Section~\ref{sec:related-work}, and then in Section~\ref{sec:definition} we describe how the problem can be defined in terms of binary relations. Section~\ref{sec:method} describes our similarity metric learning method in detail. In Section~\ref{sec:experiments} we empirically demonstrate that our method converges considerably faster than other conventional methods and that the final accuracy is higher on both intrinsic and extrinsic evaluation tasks.

\section{Related Work}
\label{sec:related-work}
Metric learning is a well established area, and much research has been put into developing sophisticated algorithms that can learn instance similarity. Metric learning techniques relevant to our work can be roughly divided into two categories: Mahalanobis based methods, and neural network based methods. Notable work in both of these areas are described, as the neural network approaches are often generalisations of the linear Mahalanobis methods to nonlinear models.

\subsection{Mahalanobis Based Methods}
The general form of a Mahalanobis distance metric parameterised by the matrix $\textbf{M}$ and defined over the set $X$ is given in Equation~\ref{eq:mahalanobis}, where $\vec x_i, \vec x_j \in X$. The algorithms based on this model primarily differ on how the linear transform is optimised.

\begin{equation}
	D(\vec x_i, \vec x_j) = \sqrt{(\vec x_i - \vec x_j)^{\top} \textbf{M} (\vec x_i - \vec x_j)}
	\label{eq:mahalanobis}
\end{equation}

The large margin nearest neighbours (LMNN)~\cite{weinberger2005} algorithm employs semidefinite programming to optimise a loss function composed of two terms. Specifically, there are two evaluations of Equation~\ref{eq:mahalanobis}. One term draws very similar instances together, and the other encourages a margin to be formed between dissimilar instances. As the name suggests, the motivation for the development of this algorithm was to improve the accuracy of $k$-Nearest Neighbours ($k$-NN) classifiers.

Information-theoretic metric learning is another Mahalanobis technique, and was introduced by~\cite{davis2007}. This criterion aims to minimise the Kullback-Leibler divergence between two multivariate Gaussians, of which the inverse covariance matrices are used to define Mahalanobis distance metrics. One of these Gaussians is defined in advance and acts as a regulariser, while the other is treated as a free parameter and optimised subject to constraints derived from similarity information.

Neighbourhood Components Analysis (NCA)~\cite{goldberger2004} is another method developed to be used in conjunction with $k$-NN classifiers. This technique attempts to find the matrix, $\textbf{M}$, by minimising a differentiable loss function that approximates the behaviour of $k$-NN in the transformed feature space.

\subsection{Neural Network Based Methods}
The first application of neural networks to metric learning was with the introduction of Siamese networks~\cite{bromley1993}. The original authors initially applied these models to signature verification, however others have since used this technique for many other domains such as face recognition and verification~\cite{taigman2014}. Siamese networks are composed of two standard feed forward neural networks that have the same topology and share the same parameters. The output of these subnetworks is then compared with the cosine similarity function to produce the final output of the network, indicating whether the instances propagated through the two subnetworks are similar or not. During training the network is presented with pairs of instances labelled either positive or negative. For positive pairs the cosine similarity is maximised, whereas for negative pairs it is minimised.

Following on from this, \cite{chopra2005} developed a variant of Siamese networks that compares the output of the subnetworks using Euclidean distance. This method was then further improved by~\cite{hadsell2006}, resulting in the contrastive loss function for Siamese networks, as given in Equation~\ref{eq:contrastive}. This function is then averaged for all training pairs to create an objective function for training the network.

\begin{equation}
\label{eq:contrastive}
L(\vec x_i, \vec x_j, y) = y \|f(\vec x_i) - f(\vec x_j)\|_2^2 + (1 - y) max(0, m - \|f(\vec x_i) - f(\vec x_j)\|_2)^2
\end{equation}

Where $\vec x_i$ and $\vec x_j$ are two instances, $y$ is the label indicating whether they are similar or dissimilar, $f$ performs a forward propagation through one of the subnetworks, and $m$ is a margin. The value of $m$ simply sets the scale of the resulting vector space and is typically set to 1.

A further generalisation of Siamese networks are the class of methods that consider three instances per loss function evaluation, so called ``triplet'' loss functions~\cite{gomez2015,schroff2015}. These methods attempt to minimise the distance between a target instance and another positive example, while simultaneously maximising the distance between the same target instance and a negative example. Some variants of this approach allow one to define the ground truth labels in terms of relative similarity. That is, rather than having hard constraints specifying similarity or dissimilarity, similarity is defined as an instance being more similar to one instance than another.

There is also an extension of NCA to nonlinear transformations of the input data~\cite{salakhutdinov2007}. This method can be viewed as a probabilistic variant of Siamese networks. The nonlinear transformation models used in the original exposition of this method were stacked Restricted Boltzmann Machines, initialised using unsupervised pretraining and subsequently fine-tuned using the NCA loss function.

A common theme that unifies all the approaches described thus far is the need to train on pairs (or triples) of instances. This increases the effective size of the training set quadratically, greatly slowing down the training time. Our proposal to decouple the process of learning the embeddings from learning the functions that performs the embedding is not entirely new. The work of~\cite{lin2014} utilises a similar two step process for learning a hashing function. In their work the embeddings are in fact bit strings, and the function used to generate the hash codes takes the form of boosted decision trees. They also use a greedy discrete optimisation procedure to find the target hash codes, rather than a numerical optimisation method to find real valued vectors.

\section{Problem Definition}
\label{sec:definition}
Similarity metric learning algorithms are typically trained on pairs of objects labelled as being either positive, for pairs of similar objects, or negative, for pairs of dissimilar objects. More formally, we have a set of objects that we call $X$. We also have a set, $Z \subset X \times X \times \{1, 0\}$, that contains pairs of objects from $X$ coupled with labels indicating whether they are similar or not. One can say that $Z$ represents a binary relation where some entries of the relation matrix are unknown.

This is a convenient problem formulation that naturally gives rise to models that are trained on pairs of objects at a time. The problem with this approach lies in the efficiency of pairwise training. The training set size is effectively squared, resulting in a representation of the training data that does not efficiently encode the useful information required to construct an accurate model. Even though all the information is available to the model, this inefficient encoding significantly slows down the training procedure.

One must inevitably train on pairs at some point in the metric learning process, however the goal of our work is demonstrate a method for modifying pre-existing loss functions in such a way that the pairwise training comprises a negligible fraction of the runtime.

A key advantage of posing similarity learning as the task of inferring a binary relation is the ability decompose the training data into classes. This idea is most often studied in conjunction with equivalence relations, where the equivalence classes form a disjoint partition over the original training set. In this context the relation must exhibit reflexivity, symmetry, and transitivity---the last of which is somewhat limiting. One can drop the requirement for transitivity by instead considering binary tolerance relations, which are only reflexive and symmetric. As with equivalence relations, it is possible to decompose the training data into a set of classes (termed tolerance classes), however these classes need not be disjoint.

Rather than partitioning the training set into, potentially overlapping, subsets that each correspond to a tolerance class, we find a target vector for each instance. These vectors are constrained such that the targets of instances that are related will be close in the target vector space and unrelated instances will be far apart. The rationale behind this is that each tolerance class will be mapped to a cluster in the target vector space, however we can employ a technique that does not have to explicitly determine to which tolerance classes each instance belongs.

\section{Method}
\label{sec:method}
Instead of training a large Siamese network on pairs of---potentially quite large---feature vectors in the training set, we compute a target vector for each training instance. To do this, we completely disregard any information provided by the features associated with each instance and instead try to solve an optimisation problem that encourages the target vectors of similar instances to be clustered together. After the target vectors have been computed, a multi-target regression model can be trained to embed instances into the target vector space. Provided that a suitable loss function has been chosen for learning the target vectors, some predefined distance metric applied to the target vector space will result in a system capable of determining instance similarity. The assumption that underlies this method is that the confusability between the latent classes in the dataset does not provide information that is useful for constructing a metric.

\subsection{Learning Target Vectors}
We now describe the method more formally, but still with enough abstraction that the generality is obvious. Consider $L(\vec x_i, \vec x_j, y_{ij})$, a differentiable loss function over a pair of instances. Similarity metric learning algorithms that rely on embedding data into a space where the semantic similarity is more salient usually rely on several components: the objective function, the model, and a fixed distance such as Euclidean or Manhattan distance. The use of a particular objective function generally implies that a certain fixed distance metric should be used on the embedded data. For example, when using the contrastive loss given in Equation~\ref{eq:contrastive} it is fairly obvious that Euclidean distance (or squared Euclidean distance) is the intended metric. This leaves two components; the objective function and the embedding model. For this class of metric learning algorithms, where an embedding function is a primary component, one can write the loss function as $L(f_\Theta(\vec x_i), f_\Theta(\vec x_j), y_{ij})$, where $f_\Theta$ is an embedding function parameterised by $\Theta$.

Consider the scenario where $f_\Theta$ is modelled as a lookup table, so each feature vector, $\vec x_i$, is simply mapped to a response vector, $t_i$, contained within $\Theta$. One can then solve the following optimisation problem:

\begin{equation}
	\Theta^\ast = \underset{\Theta}{\operatorname{arg}\,\operatorname{min}} \frac{1}{\|Z\|} \sum_{(\vec x_i, \vec x_j, y_{ij}) \in Z} L(f_\Theta(\vec x_i), f_\Theta(\vec x_j), y_{ij})
\end{equation}

Because we only consider scenarios where $L$ is differentiable, this problem could be solved with any of a large variety of numerical optimisation algorithms.

It is trivial to see why systems such as this are not used in practice, because as soon as a novel feature vector is encountered where the label is unknown the model is not capable of making a prediction. However, we can take each of the learned target vectors and train a second model, $g_\Omega$, that implements a more generalisable model. The new model can be created with any multi-target regression algorithm, but we focus on the use of deep neural networks in this paper. Although the original model for $f_\Theta$ must be trained on pairs of instances, due to the composition with $L$, this new model does not require pairwise training.

We investigate two options for $L$: the contrastive loss given in Equation~\ref{eq:contrastive}, and the loss function we define in Equation~\ref{eq:dotloss}:

\begin{equation}
	\label{eq:dotloss}
	L(f_\Theta(\vec x_i), f_\Theta(\vec x_j), y_{ij}) = \frac{1}{2} (y_{ij} - f_\Theta(\vec x_i) \cdot f_\Theta(\vec x_j))^2
\end{equation}

Where $\cdot$ represents the dot product between two vectors. Using this loss function is equivalent to factoring the adjacency matrix of the underlying binary relation defined on the training set. We use the Adam optimiser~\cite{kingma2014} to minimise both of these loss functions in a matter of seconds for each dataset we consider in our empirical evaluation.


\subsection{Learning an Embedding Function}
Once the target vectors have been found a regression model can be trained to minimise the squared error between the embedding of each instance and the corresponding target vector, as shown in Equation~\ref{eq:squared-error}, where $g_\Omega$ is the multi-target regression model. Our technique does not rely on a specific regression algorithm, but is instead very general. It is possible to use any multi-target regression method to learn a mapping from features to the learned target vectors, however our focus is on the performance of neural networks for metric learning and hence that is our regression algorithm of choice.

\begin{equation}
	\label{eq:squared-error}
	\Omega^\ast = \underset{\Omega}{\operatorname{arg}\,\operatorname{min}} \frac{1}{2\|X\|} \sum_{\vec x_i \in X} (f_{\Theta^\ast}(\vec x_i) - g_{\Omega}(\vec x_i))^2
\end{equation}

The original Siamese network introduced by~\cite{bromley1993} applied the cosine similarity function to embeddings of instances, as computed by the branches of the network, in order to determine whether instances are related. In this case a value of one means the instances are very similar, and a value of negative one indicates the instances are very dissimilar. Networks trained with the contrastive loss replace the cosine similarity function with Euclidean distance and the interpretation of the resulting real value is also changed. A value of zero indicates a high degree of similarity, and as the value becomes larger the instances are considered increasingly dissimilar.

The target vectors learned using the technique presented herein are found by either optimising loss function based on the squared Euclidean distance or the dot product. When using Euclidean distance, small values indicate similar instances and large values indicate dissimilar instances. It is important that the correct fixed metric is used on the resulting embeddings to ensure the optimal performance.

Ultimately, for all of these methods a threshold must be chosen if the problem is to be reduced to answering the question of whether two instances should be considered similar.

\section{Experiments}
\label{sec:experiments}
The motivation behind introducing this method was to accelerate the training process of neural network based metric learning algorithms. Firstly, we show that the optimisation problem used to compute the target vectors can be solved in a matter of seconds, thus comprising a negligible fraction of the overall training time. Then we demonstrate the time taken for our method and Siamese networks to converge when trained on the same datasets. Finally, we perform an extrinsic evaluation to show that the learned metrics perform well on $k$-NN classification tasks.

\subsection{Datasets and Network Architectures}
Standard image classification datasets, summarised in Table~\ref{tbl:datasets}, are used to demonstrate the capabilities of our method. The similarity metric learning methods we consider all involve pairwise training at some point, which necessitates datasets that contain pairs of instances with binary similarity constraints. In other words, for each dataset we must define a binary relation represented in the same manner as described in Section~\ref{sec:definition}. For each element in each of the datasets, 10 positive and 10 negative pairs are generated. The pairs are labelled positive if the two instances have at least one class in common. This process is performed separately with the training and testing instances to prevent overlap between the train and test subsets. 

The two most basic datasets considered are MNIST~\cite{lecun1998} and CIFAR-10~\cite{krizhevsky2009}. Both consist of 10 balanced classes and a similar number of total instances. The primary difference is that MNIST contains very easily discriminated hand written digits, and CIFAR-10 contains downsampled photographs of natural objects.

The Public Figures dataset~\cite{kumar2009} is a large collection of photos spanning 200 different identities. Unfortunately the originators of this dataset only supply URLs for the images, and because the dataset is now several years old many of these links are dead. Fortunately there is a subset called PubFig83 that has been scraped and made available for download by~\cite{pinto2011}. We use a version of this PubFig83 dataset created by~\cite{chiachia2014} that has had the faces aligned such that the eyes in each image are always in the same position. In this version of the dataset there are 13,838 colour images that are all $100\times100$ pixels. The networks trained on this dataset are only supplied with the central $60\times60$ pixels of each image in order to reduce overfitting caused by the background clutter surrounding the faces.

NUS-WIDE~\cite{chua2009}, the multi-label dataset, is included in order to simulate a tolerance relation. Although there are only 81 classes, this dataset includes 16,458 unique label vectors consisting of different combinations of these classes. This results in a highly complex metric learning problem. The difficulty of this dataset is further compounded by the presence of label noise, due to the labels being determined using a method that relies on user specified tags. The original dataset consists of a set of image URLs and associated labels, however some of these URLs are now unavailable. We managed to collect 222,654 of the original 269,648 instances. Each image was resized such that the smallest dimension was 100, and then the central $100 \times 100$ pixels were used for training and testing.

\begin{table}
	\center
	\caption{A summary of the datasets used for the experiments in this paper.}
	\label{tbl:datasets}
	\begin{tabular*}{0.9\columnwidth}{@{\extracolsep{\fill}}ccccc}
		\hline\noalign{\smallskip}
		Dataset & Train Instances & Test Instances & Features & Labels \\
		\noalign{\smallskip}
		\hline
		\noalign{\smallskip}
		MNIST & 60,000 & 10,000 & 784 & 10 \\
		CIFAR-10 & 50,000 & 10,000 & 3,072 & 10 \\
		PubFig83 & 11,000 & 2,838 & 30,000 & 83 \\
		NUS-WIDE & 150,000 & 72,654 & 30,000 & 81 \\
		\noalign{\smallskip}
		\hline
	\end{tabular*}
\end{table}

Each dataset requires a different network due to the varying complexity of the associated task and the different number of features contained within each instance. Table~\ref{tbl:architectures} provides an overview of these architectures. In the case of a Siamese network the architectures given in Table~\ref{tbl:architectures} describe only a single branch of the network. Additionally, when training networks on CIFAR-10, PubFig83, and NUS-WIDE, we also train on horizontal flips of images in the training set. The size of the output layer determines the length of the learned target vectors for each dataset.

\begin{table}
	\center
	\caption{The different network architectures used for each dataset throughout all experiments. In this table Dense $x$ indicates a fully connected layer with $x$ hidden units, Convolutionall $x\times y\times z$ means a convolutional layer with $x$ feature maps and filters of size $y\times z$. Lastly, Max Pool $x \times y$, $z \times w$ represents a max pooling layer with a pool size of $x \times y$ and a stride of $z \times w$.}
	\label{tbl:architectures}
	\begin{tabular*}{0.7\columnwidth}{@{\extracolsep{\fill}}cc}
		\hline\noalign{\smallskip}
		MNIST & PubFig83 \\
		\noalign{\smallskip}
		\hline
		\noalign{\smallskip}
		Dense 500 & Convolutional $64 \times 9 \times 9$ \\
		Dense 500 & Max Pool $2 \times 2$, $2 \times 2$ \\
		Dense 16 & Convolutional $96 \times 7 \times 7$ \\
		& Max Pool $2 \times 2$, $2 \times 2$ \\
		& Convolutional $128 \times 7 \times 7$ \\
		& Dense 1,024 \\
		& Dense 1,024 \\
		& Dense 64 \\
		\noalign{\smallskip}
		\hline\noalign{\smallskip}
		CIFAR-10 & NUS-WIDE \\
		\noalign{\smallskip}
		\hline
		\noalign{\smallskip}
		Convolutional $64 \times 3 \times 3$ & Convolutional $64 \times 3 \times 3$ \\
		Convolutional $64 \times 3 \times 3$ & Convolutional $64 \times 3 \times 3$ \\
		Max Pool $3 \times 3$, $2 \times 2$ & Convolutional $64 \times 3 \times 3$ \\
		Convolutional $96 \times 3 \times 3$ & Max Pool $2 \times 2$, $2 \times 2$ \\
		Convolutional $96 \times 3 \times 3$ & Convolutional $96 \times 3 \times 3$ \\
		Max Pool $3 \times 3$, $2 \times 2$ & Convolutional $96 \times 3 \times 3$ \\
		Dense 128 & Convolutional $96 \times 3 \times 3$ \\
		Dense 128 & Max Pool $2 \times 2$, $2 \times 2$ \\
		Dense 16 & Convolutional $128 \times 3 \times 3$ \\
		& Convolutional $128 \times 3 \times 3$ \\
		& Convolutional $128 \times 3 \times 3$ \\
		& Max Pool $2 \times 2$, $2 \times 2$ \\
		& Dense 4,096 \\
		& Dense 4,096 \\
		& Dense 32 \\
		\noalign{\smallskip}
		\hline
	\end{tabular*}
\end{table}

For each hidden layer the ReLU activation function is used, and the output layers do not use any nonlinearity. Dropout~\cite{srivastava2014} (with $p=0.5$) was applied before all hidden layers consisting of fully connected units. The weight initialisation procedure of~\cite{glorot2010} was used for setting the starting values for the weights in all network. We applied a slightly modified standardisation procedure to the target vectors when training the regression networks. Mean subtraction is performed, however rather than dividing by the individual standard deviation for each component in the target vectors we scale all vectors such that the mean standard deviation of all the components is one. This prevents distortion of the target vector space, which would effectively change the objective function to a weighted squared error variant.

\subsection{Implementation}
The target vector optimisation was performed using a single threaded program written in the D programming language\footnote{http://www.dlang.org} and run on an Intel i7 4770. The deep networks are trained using an implementation that takes advantage of functions provided by cuDNN 4~\cite{chetlur2014} and is executed on an NVIDIA TITAN X GPU.

\subsection{First Phase Optimisation}
The first point of order is to show that the first optimisation problem consumes a negligible fraction of the overall training time when constructing a learned similarity metric. Table~\ref{tbl:first-phase} contains empirical estimates of the expected runtime for both loss functions when applied to each dataset. Although the dot product based loss function takes longer to finish than the contrastive loss, both still finish in a relatively short amount of time.

\begin{table}
	\center
	\caption{95\% confidence intervals for the expected runtime of the first phase optimisation problem for each loss function and dataset combination. Values are in seconds. FML-C denotes targets trained with the contrastive loss function and FML-DP denotes targets trained with the loss function given in Equation~\ref{eq:dotloss}.}
	\label{tbl:first-phase}
	\begin{tabular*}{0.9\columnwidth}{@{\extracolsep{\fill}}ccccc}
	\hline\noalign{\smallskip}
	& MNIST & CIFAR-10 & PubFig83 & NUS-WIDE \\
	\noalign{\smallskip}
	\hline
	\noalign{\smallskip}
	FML-C & 6.90 ($\pm$0.01) & 5.19 ($\pm$0.21) & 1.94 ($\pm$0.01) & 46.43 ($\pm$1.91) \\
	FML-DP & 21.77 ($\pm$1.38) & 15.89 ($\pm$1.70) & 6.86 ($\pm$0.18) & 152.30 ($\pm$4.89) \\
	\noalign{\smallskip}
	\hline
	\end{tabular*}
\end{table}

\subsection{Time to Converge}
\begin{figure}
	\begin{subfigure}{0.5\textwidth}
		\center
		\begin{tikzpicture}[scale=0.7,trim axis right,trim axis left]
			\begin{axis}[yticklabel style={/pgf/number format/fixed,
			                     /pgf/number format/precision=3}, legend pos=south east, ylabel=AUROC, xlabel={Time (Seconds)}]
				\addplot +[mark=none] table[col sep=comma] {results/mnist-fml-mf.csv};
				\addlegendentry{FML-DP}
				\addplot +[mark=none] table[col sep=comma] {results/mnist-fml-contrastive.csv};
				\addlegendentry{FML-C}
				\addplot +[mark=none] table [col sep=comma] {results/mnist-siamese-contrastive.csv};
				\addlegendentry{Siamese}
			\end{axis}
		\end{tikzpicture}
		\caption{MNIST}
	\end{subfigure}
	\begin{subfigure}{0.5\textwidth}
		\center
		\begin{tikzpicture}[scale=0.7,trim axis right,trim axis left]
			\begin{axis}[yticklabel style={/pgf/number format/fixed,
			                     /pgf/number format/precision=3}, legend pos=south east, ylabel=AUROC, xlabel={Time (Seconds)}]
				\addplot +[mark=none] table [col sep=comma] {results/cifar10-fml-mf.csv};
				\addlegendentry{FML-DP}
				\addplot +[mark=none] table [col sep=comma] {results/cifar10-fml-contrastive.csv};
				\addlegendentry{FML-C}
				\addplot +[mark=none] table [col sep=comma] {results/cifar10-siamese-contrastive.csv};
				\addlegendentry{Siamese}
			\end{axis}
		\end{tikzpicture}
		\caption{CIFAR-10}
	\end{subfigure} \\[1cm]
	\begin{subfigure}{0.5\textwidth}
		\center
		\begin{tikzpicture}[scale=0.7,trim axis right,trim axis left]
			\begin{axis}[yticklabel style={/pgf/number format/fixed,
			                     /pgf/number format/precision=3}, legend pos=south east, ylabel=AUROC, xlabel={Time (Seconds)}]
				\addplot +[mark=none] table [col sep=comma] {results/pubfig83-fml-mf.csv};
				\addlegendentry{FML-DP}
				\addplot +[mark=none] table [col sep=comma] {results/pubfig83-fml-contrastive.csv};
				\addlegendentry{FML-C}
				\addplot +[mark=none] table [col sep=comma] {results/pubfig83-siamese-contrastive.csv};
				\addlegendentry{Siamese}
			\end{axis}
		\end{tikzpicture}
		\caption{PubFig83}
	\end{subfigure}
	\begin{subfigure}{0.5\textwidth}
		\center
		\begin{tikzpicture}[scale=0.7,trim axis right,trim axis left]
			\begin{axis}[yticklabel style={/pgf/number format/fixed,
			                     /pgf/number format/precision=3},
								 scaled ticks=false, tick label style={/pgf/number format/fixed}, legend pos=south east, ylabel=AUROC, xlabel={Time (Seconds)}]
				\addplot +[mark=none] table [col sep=comma] {results/nuswide-fml-mf.csv};
				\addlegendentry{FML-DP}
				\addplot +[mark=none] table [col sep=comma] {results/nuswide-fml-contrastive.csv};
				\addlegendentry{FML-C}
				\addplot +[mark=none] table [col sep=comma] {results/nuswide-siamese-contrastive.csv};
				\addlegendentry{Siamese}
			\end{axis}
		\end{tikzpicture}
		\caption{NUS-WIDE}
	\end{subfigure}
	\caption{Plots of the AUROC on the test set vs training time for each dataset. In these plots FML denotes one of the fast metric learning methods presented in this work. The suffix DP means the targets were found with the loss function given in Equation~\ref{eq:dotloss}, and C means they were found with the contrastive loss. Each Siamese network was run until convergence, and then the other methods were run for double the number of epochs taken to train the Siamese network. Because the Siamese network considers pairs of images, each epoch takes twice as long as a regular regression network. On the NUS-WIDE experiment we stopped the FML methods earlier due to time constraints and the lack of extra information that would be obtained from running them for the same duration as the Siamese network.}
	\label{fig:convergence}
\end{figure}
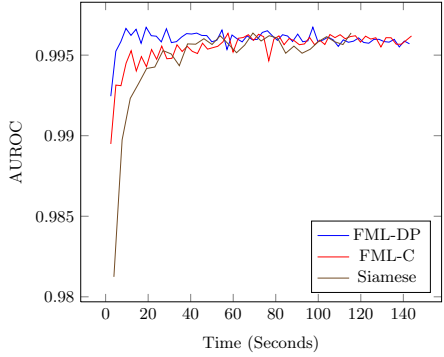
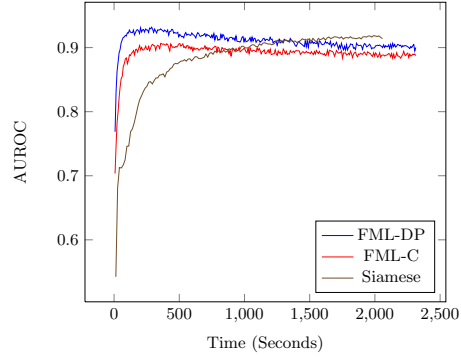
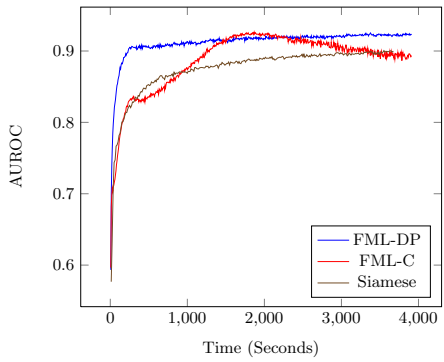
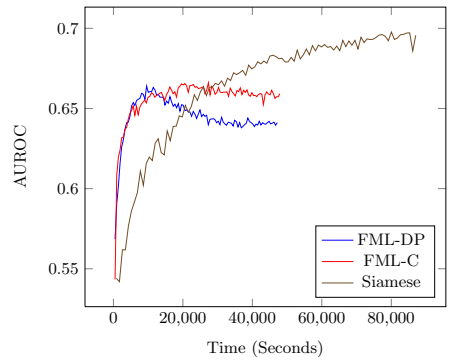

We now demonstrate that our method converges significantly faster than a conventional Siamese network trained with the contrastive loss function, while still achieving competitive accuracy on an intrinsic evaluation task. Because the range of the various loss functions used to train the different models we are evaluating are quite different we must select a single metric to use as a proxy for model performance. We have chosen the Area Under the Receiver Operating Characteristic curve (AUROC), where the binary classification task is to determine whether two instances are similar or not according to the previously defined binary relation. Figure~\ref{fig:convergence} shows how fast the three methods converge for each dataset. 

We can see that both variants of the fast metric learning method that has been proposed converge significantly faster than the conventional Siamese network. However, on one dataset the Siamese network does outperform our method. 

\subsection{$k$-NN Classifier Performance}
To perform an extrinsic evaluation that shows how useful this method can be in practice, $k$-NN classifiers are created that utilise Euclidean distance applied to the embeddings to make predictions. A validation set was used for determining how many epochs each network should be trained for. It should be noted that the models trained with the dot product based loss function given in Equation~\ref{eq:dotloss} are at a disadvantage in this case. The multiclass classifiers were trained using WEKA~\cite{hall2009}, and the multi-label classifiers are created using the binary relevance problem transformation scheme as implemented in MEKA.\footnote{http://meka.sourceforge.net} Table~\ref{tbl:knn} shows the performance of the $k$-NN classifiers, with $k = 5$ for all classifiers.

\begin{table}
	\center
	\caption{Performance of $k$-NN classifiers applied to each dataset and model combination. Accuracy is reported for the three multiclass datasets (MNIST, CIFAR-10, and PubFig83), and the Jaccard index (higher is better) is reported for the multi-label dataset (NUS-WIDE).}
	\label{tbl:knn}
	\begin{tabular*}{0.9\columnwidth}{@{\extracolsep{\fill}}ccccc}
		\hline\noalign{\smallskip}
		Algorithm & MNIST & CIFAR-10 & PubFig83 & NUS-WIDE \\
		\noalign{\smallskip}
		\hline
		\noalign{\smallskip}
		FML-C & 0.976 & 0.806 & 0.777 & 0.427 \\
		FML-DP & 0.978 & 0.806 & 0.768 & 0.229 \\
		Siamese & 0.972 & 0.734 & 0.335 & 0.393 \\
		\noalign{\smallskip}
		\hline
	\end{tabular*}
\end{table}

It can be seen that both the fast metric learning methods presented in this work outperform the conventional Siamese network trained with the contrastive loss function when evaluating on multiclass classification tasks. Particularly surprising is the performance on PubFig83, where the accuracy of the Siamese network is significantly worse than the other methods. Also interesting is the performance on the NUS-WIDE dataset. It is quite surprising that FML-C achieves a higher Jaccard index than the Siamese network, despite the intrinsic evaluation showing the Siamese network converges towards a more accurate solution.

\section{Conclusion}
In this paper, we have presented a fast method for learning similarity metrics backed by deep neural networks. It has been shown that the convergence time for the techniques presented in this work are significantly faster and, in the majority of cases, result in superior performance on both intrinsic and extrinsic evaluation tasks. Our method, coupled with the contrastive loss function, appears to be a very good choice for learning specialised distance metrics for $k$-NN.

It would be interesting to investigate how well these methods work on information retrieval tasks. The formalisation given in Section~\ref{sec:definition} is well suited for scenarios where one does not wish explicity assign a ground truth class during data collection, even though there is likely to be a large number of latent classes or topics. It would also be interesting to investigate how effective two step training phase is for speeding up networks trained with triplet loss functions, especially since these loss functions are more popular for information retrieval tasks~\cite{chechik2010}.

\subsubsection*{Acknowledgements.} We thank NVIDIA for donating the TITAN X GPU that was used for training all of the deep networks in this paper.

\bibliographystyle{plain}
\bibliography{refs}

\end{document}